\documentclass{article}
\usepackage{spconf,amsmath,amssymb,graphicx,hyperref}
\usepackage{titlesec}
\usepackage{algorithm}
\usepackage{algpseudocode}
\usepackage{amsmath,amssymb}
\usepackage{xcolor}
\usepackage{xcolor}
\usepackage{subcaption} % defines subfigure environment

\usepackage{booktabs}   % \toprule, \midrule, \bottomrule
\usepackage{multirow}   % multirow cells in tables
\usepackage{array}      % extended column definitions

\title{Cross-modal guidance for fast diffusion-based computed tomography}

\name{
\begin{tabular}{c}
Timofey Efimov$^{1,2}$, Singanallur Venkatakrishnan$^{2}$, Maliha Hossain$^{2}$, \\
Haley Duba-Sullivan$^{2,3}$, Amirkoushyar Ziabari$^{2}$
\end{tabular}
}

\address{
$^{1}$ Carnegie Mellon University, Pittsburgh, USA \\
$^{2}$ Oak Ridge National Laboratory, Oak Ridge, USA \\
$^{3}$ Purdue University, West Lafayette, USA \\[2pt]
\begin{minipage}{0.9\linewidth}\centering\small
\texttt{tefimov@andrew.cmu.edu}\\
\texttt{\{ziabariak,hossainm,sullivanhe\}@ornl.gov}\\
\texttt{ svvenkatakrishnan@gmail.com}
\end{minipage}
}
\ninept 
\begin{document}
% \ninept
\maketitle

\renewcommand{\thefootnote}{}
\footnotetext{
This manuscript has been authored by UT-Battelle, LLC, under contract DE-AC05-00OR22725 with the US Department of Energy (DOE). Research supported by an appointment to the Oak Ridge National Laboratory GRO Program, sponsored by the U.S. Department of Energy and administered by the Oak Ridge Institute for Science and Education; and through ORNL LDRD AI-initiative program.
The US government retains and the publisher, by accepting the article for publication, acknowledges that the US government retains a nonexclusive, paid-up, irrevocable, worldwide license to publish or reproduce the published form of this manuscript, or allow others to do so, for US government purposes. DOE will provide public access to these results of federally sponsored research in accordance with the DOE Public Access Plan (\url{http://energy.gov/downloads/doe-public-access-plan}).
}
\renewcommand{\thefootnote}{\arabic{footnote}}

% \end{enumerate}
\begin{abstract}
Diffusion models have emerged as powerful priors for solving inverse problems in computed tomography (CT). In certain applications, such as neutron CT, it can be expensive to collect large amounts of measurements even for a single scan, leading to sparse data sets from which it is challenging to obtain high quality reconstructions even with diffusion models.
One strategy to mitigate this challenge is to leverage a complementary, easily available imaging modality; however, such approaches typically require retraining the diffusion model with large datasets.
In this work, we propose incorporating an additional modality without retraining the diffusion prior, enabling accelerated imaging of costly modalities. 
We further examine the impact of imperfect side modalities on cross-modal guidance. 
Our method is evaluated on sparse-view neutron computed tomography, where reconstruction quality is substantially improved by incorporating X-ray computed tomography of the same samples.
\end{abstract}


% === Bring in section files ===
\section{Introduction}
\label{sec:intro}

Computational imaging often requires solving ill-posed inverse problems, where recovering an unknown signal $x^\star \in \mathbb{R}^d$ from incomplete or noisy measurements $y \in \mathbb{R}^m$ demands strong regularization. We consider the standard linear inverse problem:
\begin{equation}
    y \;=\; Ax^{\star} + \xi,
\end{equation}
where $A \in \mathbb{R}^{m \times d}$ is the forward projection operator and $\xi$ denotes measurement noise. Since $m \ll d$ or $A$ is ill-conditioned, recovering $x^\star$ is highly ill-posed, motivating the need for powerful priors. Diffusion models~\cite{ho2020denoising, song2019score, nichol2021improved} have emerged as state-of-the-art generative priors for such problems. By capturing complex high-dimensional distributions, diffusion priors achieve impressive results for inverse problems~\cite{chung2023solving, chung2024deep, kawar2022denoising, xu2024provably}, enabling reconstructions from fewer measurements, especially in imaging applications where acquisitions are costly and time-consuming. This is particularly beneficial for expensive computed tomography modalities such as neutron CT (NCT), where reducing the number of views can substantially lower acquisition time and cost~\cite{venkatakrishnan2021convolutional}. Nevertheless, even strong priors cannot fully overcome the fundamental limits imposed by physics and sampling, often leading to a loss of fine structural detail.

A natural strategy is to exploit complementary, low-cost modalities that capture additional structural information. 
In materials science, \textbf{neutron CT (NCT)} and \textbf{X-ray CT (XCT)} exemplify this complementarity~\cite{kim2013high,vlassenbroeck2007comparative,gleason2010x}: NCT is highly sensitive to light elements (e.g., hydrogen), whereas XCT emphasizes electron density. 
Because each modality reveals unique features, inexpensive XCT scans of the same specimen can provide valuable auxiliary guidance for improving ultra-sparse NCT reconstructions.
Existing cross-modal diffusion approaches~\cite{efimov2025leveraging, author202Xconversion, xia2025diffusion, chung2023prompt, li2025cross} in the context of other computational imaging applications typically embed auxiliary guidance directly into the prior, requiring modality-specific retraining — a process that is data-intensive, computationally costly, and prone to poor generalization.

In this paper, we leverage general diffusion priors trained on geometric structures, enabling application across modalities without retraining. 
To address these limitations in reconstructing high-fidelity data from sparse measurements, we introduce a lightweight \textit{cross-modal consistency network} that refines reconstructions at test time using auxiliary XCT data. 
This network enforces consistency between the current estimate from the expensive modality and the auxiliary observations, providing efficient cross-modal guidance while preserving the generality of pretrained priors. Within this framework, we show that auxiliary guidance improves reconstructions even when the guidance modality is degraded by noise, blur, or sparse sampling, and that cross-modal consistency accelerates and stabilizes test-time adaptation of diffusion priors. Finally, to support this line of research, we contribute the first dataset of registered NCT and XCT scans under diverse acquisition settings.

\section{Background}
\label{sec:background}

\subsection{Diffusion Models for Inverse Problems}
Diffusion models learn to sample from data distributions by estimating score functions across multiple noise levels via denoising score matching. 
For inverse problems, reconstruction algorithms typically alternate between (i) applying the unconditional diffusion prior and (ii) enforcing data consistency with $y$. 
This iterative process balances prior information and measurement fidelity, without task-specific retraining. 
Algorithms following this framework include DPS~\cite{chung2023solving}, DPnP~\cite{xu2024provably}, FPS-SMC for linear inverse problems~\cite{dou2024diffusion}, and D3IP~\cite{chung2024deep}. A key challenge is distribution shift between training and test data, as in CT with mismatched noise levels or structural characteristics. 
The D3IP algorithm~\cite{chung2024deep} addresses this issue through \textit{test-time fine-tuning}, where priors trained on general geometric shapes are adapted for specific domains such as MRI and CT. 

\subsection{Cross-modal inverse problems with deep priors}
Recent work has explored incorporating auxiliary modalities into deep prior-based reconstruction through joint training or cross-modal conditioning~\cite{efimov2025leveraging, author202Xconversion, xia2025diffusion, chung2023prompt, li2025cross}. 
However, such methods require retraining for each modality pair and often assume overly simplified corruption models for the auxiliary data. 
In contrast, our approach leverages general-purpose diffusion priors trained on universal microstructure representations, combined with a lightweight cross-modal consistency module. 
This design eliminates the need for retraining, accommodates arbitrary auxiliary degradations, and seamlessly extends the unimodal approach to cross-modal setting.

% !TEX root = ../main.tex
\section{Algorithm}
\label{sec:algorithm}

\newcommand{\Cset}[1]{\mathcal{C}\!\left(#1\right)}
\newcommand{\Taux}{\mathcal{T}_{\phi}}

\newcommand{\ProjC}[2]{\mathrm{Proj}_{\Cset{#1}}\!\left(#2\right)}

\begin{figure}[t]
  \centering
  \includegraphics[width=\columnwidth]{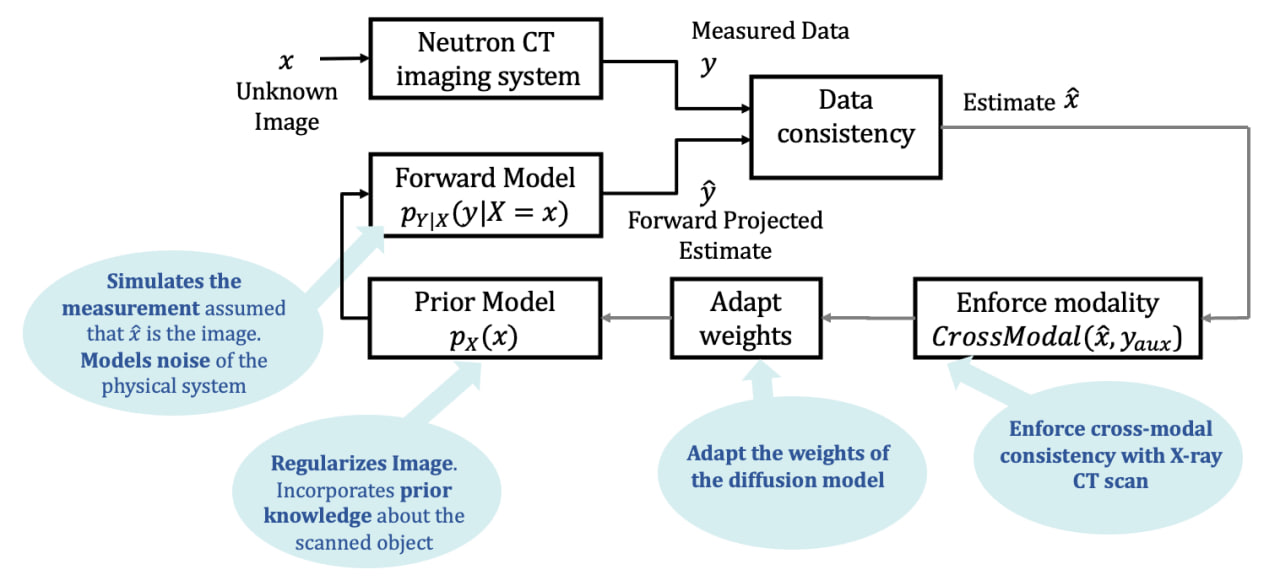}
  \caption{The cross-modal approach decouples fitting main modality measurements from enforcing cross-modal consistency. In the first step, we fine-tune the diffusion model weights to better fit the data consistency loss, then obtain the reconstruction estimate via any diffusion-based inverse problem solver. After that stage, cross-modal consistency is enforced via a lightweight image translation model. The cross-modal consistency block can be seamlessly incorporated as a separate module without modifying the prior.}
  \label{fig:cross-modal}
\end{figure}

\begin{algorithm}[t]
\caption{Cross-modal Out-of-Distribution diffusion inverse problem solver (Adapted from D3IP~\cite{chung2024deep})} 
\label{alg:mc-ood-diff}
\begin{algorithmic}[1]
\Require $\theta$ (general diffusion prior for microstructures), $N$, $T'$, $K$, $\eta$, $\lambda$, $y_{\text{main}}$ (observed main modality), $y_{\text{aux}}$ (observed auxiliary modality), $A$ (forward model for main modality), \texttt{Num\_steps}, learning rate $\gamma$
\State $\mathcal{L}(x, y_{\text{main}}, \mathrm{DiffSolver}_{\theta}) \gets \left\lVert y_{\text{main}} - A\, \mathrm{DiffSolver}_{\theta}(x \mid y_{\text{main}}) \right\rVert_2^2$ 
\State $\theta_{T'} \gets \theta$

\State $X_{T'} \gets \sqrt{\bar{\alpha}_{T'}}\, A^{\dagger} y_{\text{main}} \;+\; \sqrt{1-\bar{\alpha}_{T'}}\, \epsilon$, $\epsilon \sim \mathcal{N}(0,I)$ 
\For{$t = T'$ \textbf{to} $1$}
  \State $\theta_{t}^{(0)} \gets \theta_{t}$ 
  \For{$i = 1$ \textbf{to} \texttt{Num\_steps}} 
    \State $x_{t}^{\{i\}},\, y_{\text{main}}^{\{i\}} \sim \mathrm{MC}\!\left((X_{t}, y_{\text{main}}),\, K\right)$
    \State $g \gets \nabla_{\theta} \mathcal{L}\!\left(x_{t}^{\{i\}},\, y_{\text{main}}^{\{i\}},\, \mathrm{DiffSolver}^{t}_{\theta}\right)\Big|_{\theta=\theta_{t}^{(i-1)}}$
    \State $\theta_{t}^{(i)} \gets \theta_{t}^{(i-1)} - \gamma \, g$
  \EndFor
  \State $\theta_{t-1} \gets \theta_{t}^{(\texttt{Num\_steps})}$ 
  \State $\hat{X}_{0\mid t} \gets \mathrm{DiffSolver}^{t}_{\theta_{t-1}}\!\left(X_{t} \mid y_{\text{main}}\right)$

\If{$t \bmod 2 = 0$ \textbf{and} $t > 1$}
  \State $\tilde{X}_{0\mid t} \gets 
     \colorbox{yellow!20}{$\text{Cross-modalModel}\!\left(\hat{X}_{0\mid t}, y_{\text{aux}}\right)$}$
\Else
  \State $\tilde{X}_{0\mid t} \gets \hat{X}_{0\mid t}$
\EndIf

  \State $X_{t-1} \gets \sqrt{\bar{\alpha}_{t-1}}\, \tilde{X}_{0\mid t} \;+\; \sqrt{1-\bar{\alpha}_{t-1}}\, \epsilon$, $\epsilon \sim \mathcal{N}(0,I)$
\EndFor
\State \textbf{return} $X_{0}$
\end{algorithmic}
\end{algorithm}

The key distinction of our approach is the explicit decoupling of the diffusion prior from the cross-modal consistency mechanism. 
Fig.~\ref{fig:cross-modal} describes the overall block diagram of the approach. 
Here, we first apply a unimodal diffusion prior for the target modality and enforce data consistency with the neutron CT projections alone, without involving the cross-modal guidance. Only afterward do we pass the estimate through a lightweight cross-modal consistency module, which aligns it with the auxiliary X-ray image. This separation allows us to avoid retraining the diffusion prior with the cross-modal guidance, and instead separate the guidance with an easily trained network. We then apply the domain adaptation step to steer the diffusion weights to better capture the true probability distribution of images.  The key difference between our approach and D3IP~\cite{chung2024deep} is in the cross-modal consistency step, which enforces the cross-modal alignment for the predicted NCT reconstruction at each step of the algorithm. 

In Alg.~\ref{alg:mc-ood-diff}, we denote by $\theta$ the parameters of a general diffusion prior trained on ellipses, resembling microstructures, following the D3IP work~\cite{chung2024deep}. In D3IP, given a forward model $A$ for the main modality, we use a diffusion-based inverse problem solver, denoted as $\mathrm{DiffSolver}_{\theta}(\cdot \mid y_{\text{main}})$, which generates reconstruction estimates conditioned on the observed data. Then the domain adaptation step of D3IP fine-tunes the prior parameters $\theta$ toward the target distribution by minimizing the data-consistency loss
\begin{equation}
\mathcal{L}(x, y_{\text{main}}, \mathrm{DiffSolver}_{\theta}) 
= \| y_{\text{main}} - A\, \mathrm{DiffSolver}_{\theta}(x \mid y_{\text{main}})\|_2^2,
\end{equation}
ensuring that the prior remains faithful to the measurements. At each reverse diffusion step $t$, the algorithm maintains a noisy latent $X_t$ and adapted parameters $\theta_t$. The key mechanism of D3IP is the alternation between \emph{domain adaptation} and \emph{prediction}: given $(X_t,\theta_t)$, the weights are first updated via gradient descent on the loss
\begin{equation}
\theta_t \;\leftarrow\; \theta_t - \gamma \, \nabla_{\theta_t}\|y_{\text{main}} - A\,\mathrm{DiffSolver}_{\theta_t}(X_t \mid y_{\text{main}})\|_2^2,
\end{equation}

and immediately thereafter, the adapted solver is used to predict a reconstruction
\begin{equation}
\hat{X}_{0|t} = \mathrm{DiffSolver}_{\theta_t}^t(X_t \mid y_{\text{main}}),
\end{equation}
This alternating structure gradually improves reconstruction quality while steering the prior parameters toward the target domain. To handle a large amount of 3D data, subsets of slices from the 3D volume are sampled at each inner iteration, enabling efficient adaptation without requiring the entire volume at once. Cross-modal consistency is enforced only every other iteration to get more information from the measurements and also reduce the computational expense introduced by running the additional network.

Our key contribution is a cross-modal refinement module that integrates auxiliary XCT data into D3IP without retraining the diffusion prior. After obtaining $\hat{X}_{0|t}$, we refine it with a cross-modal network, i.e. $\text{Cross-modalModel}(\hat{X}_{0|t}, y_{\text{aux}})$ that leverages degraded auxiliary modality information (e.g., XCT) to produce $\tilde{X}_{0|t}$. 
The cross-modal network captures the shared information between NCT and XCT, and removes the artifacts in the auxiliary modality. This step not only improves the reconstruction of the NCT image but also better guides the domain adaptation, providing a better starting point for it in the early stages of the algorithm execution.

\begin{figure}[!h]
\centering
% First column
\begin{minipage}{0.15\textwidth}
    \centering
    \includegraphics[width=\textwidth]{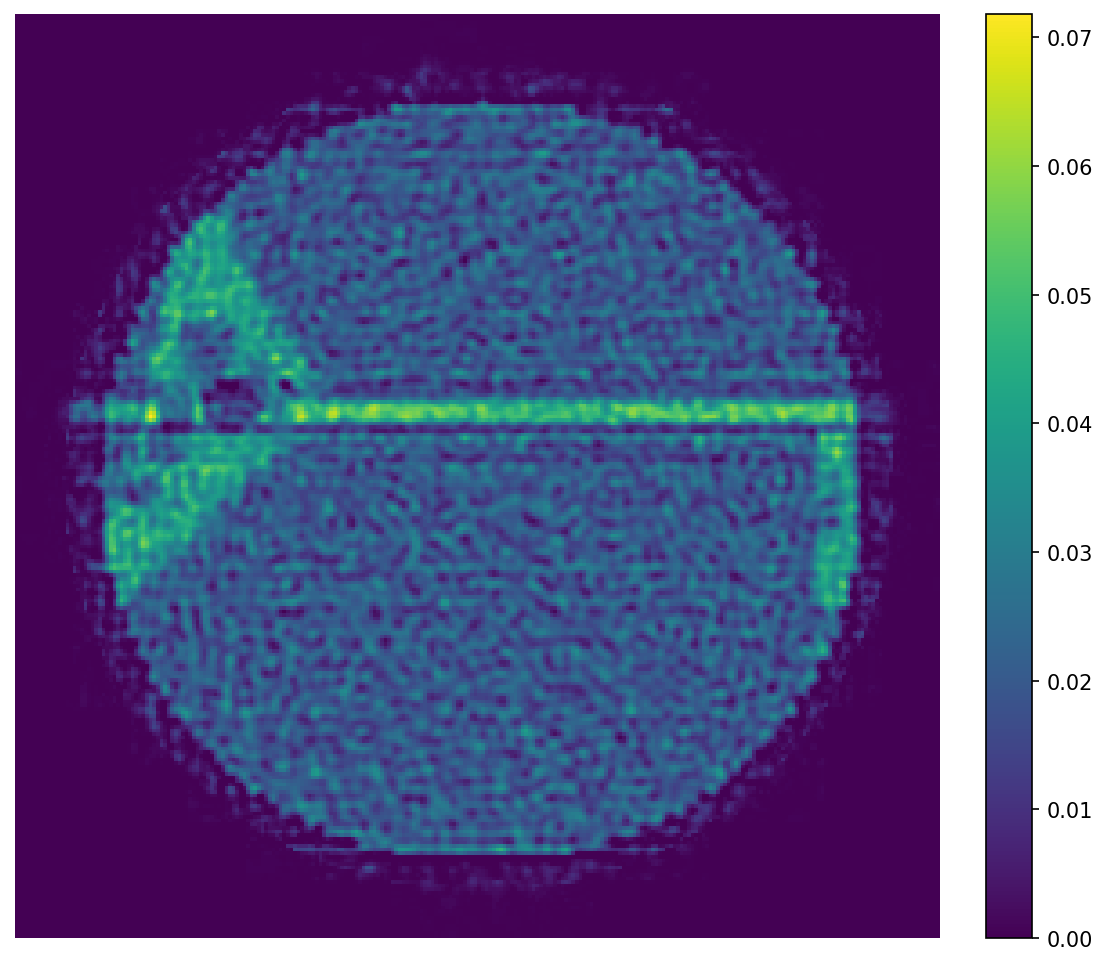}\\[2pt]
    \includegraphics[width=\textwidth]{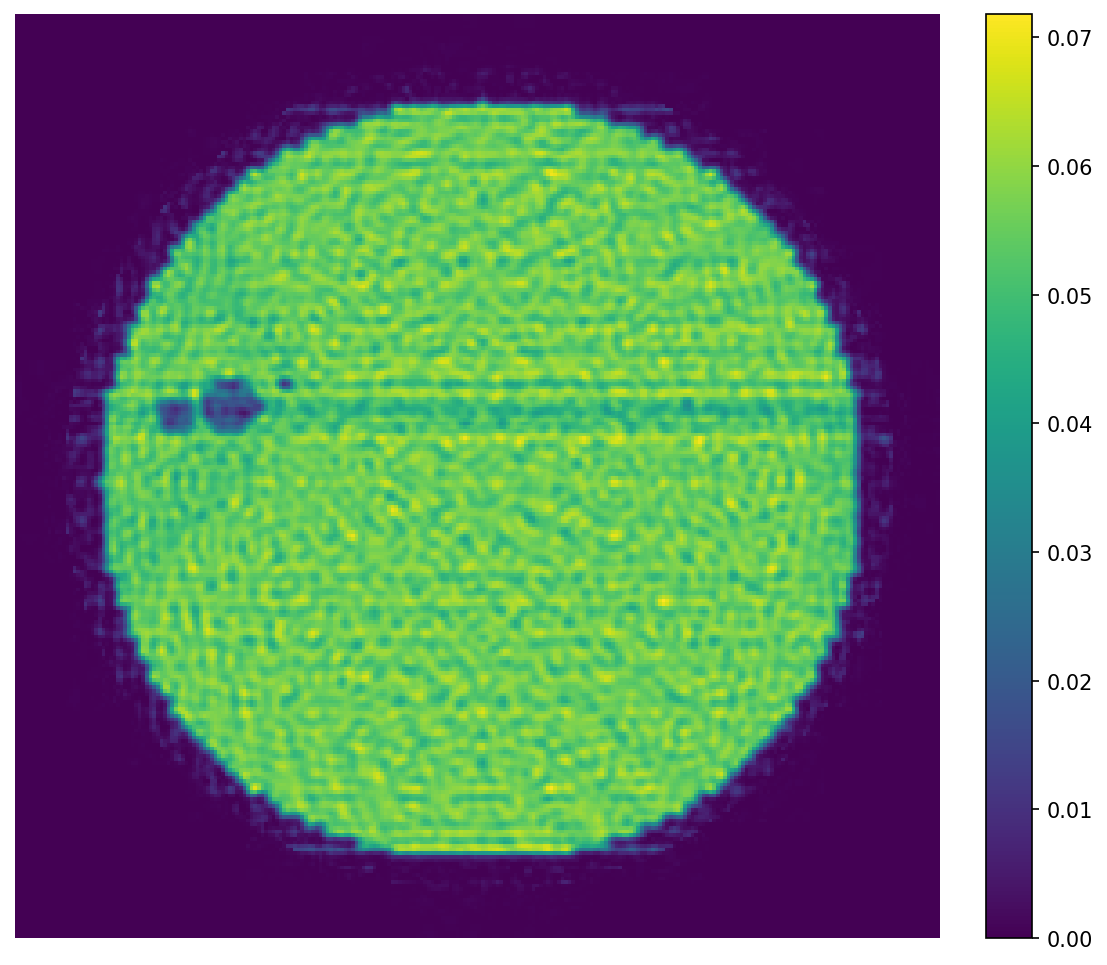}
\end{minipage}
\hspace{0.05\textwidth}
% Second column
\begin{minipage}{0.15\textwidth}
    \centering
    \includegraphics[width=\textwidth]{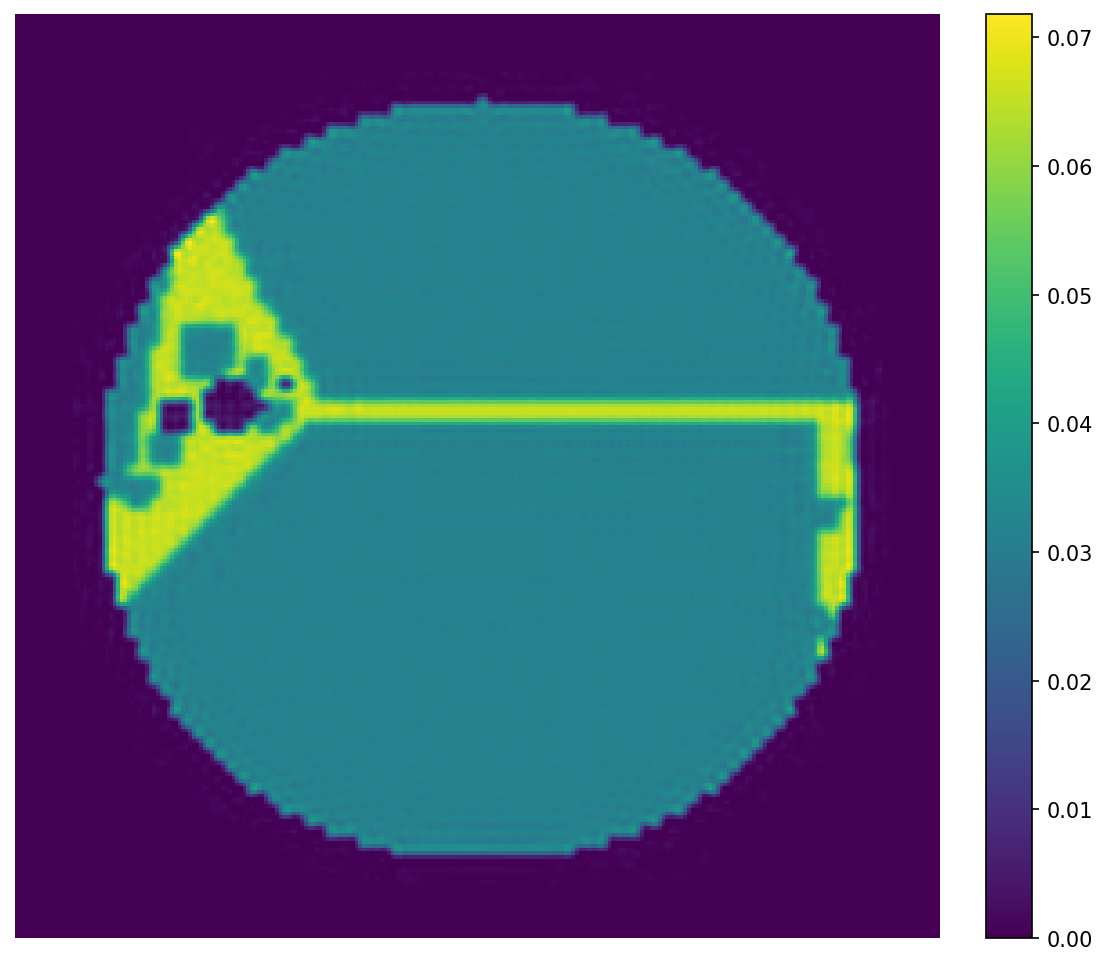}\\[2pt]
    \includegraphics[width=\textwidth]{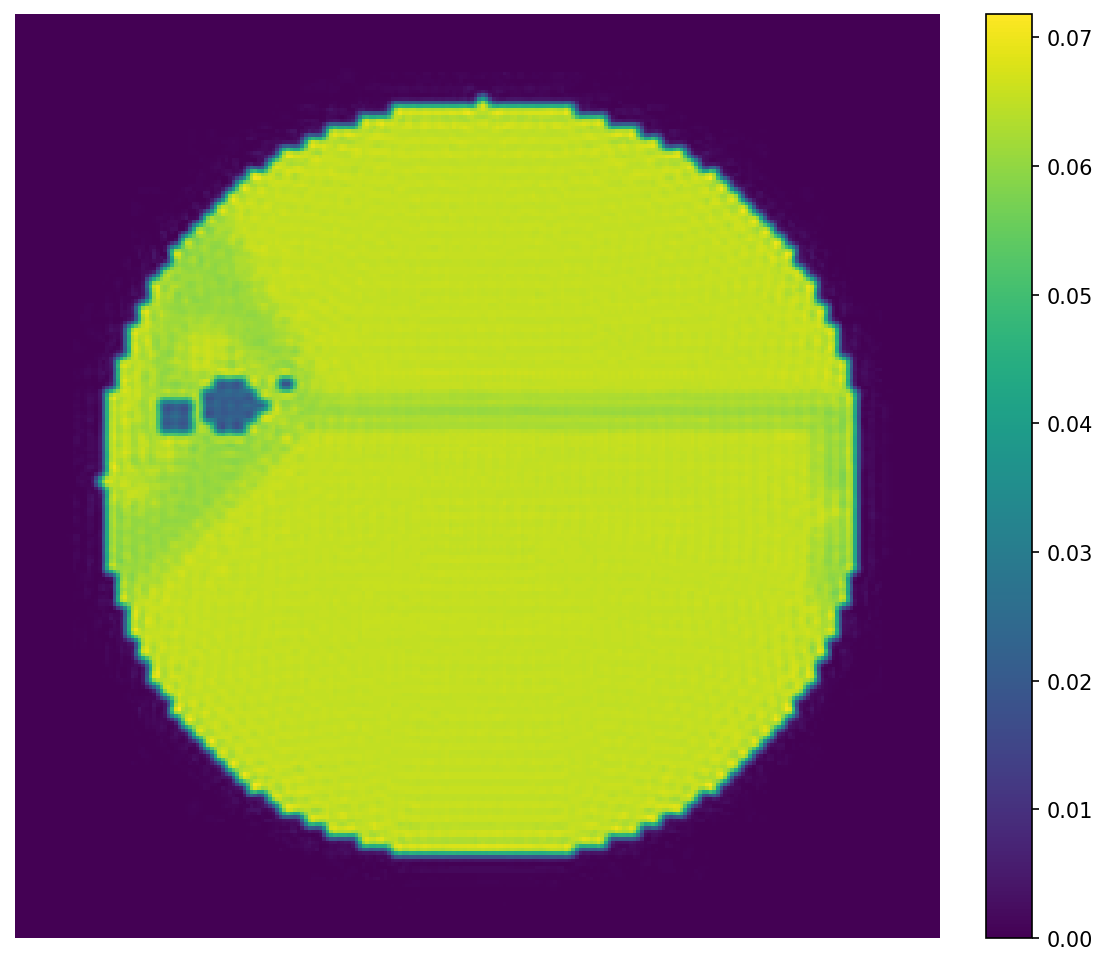}
\end{minipage}

\caption{Paired degraded NCT/XCT samples (left) and their high-quality reconstructions (right). These examples illustrate the training data used for cross-modal translation.}

\label{fig:100grid}
\end{figure}

Cross-modal model jointly leverages both the NCT estimate and the auxiliary sample that is quickly acquired using X-ray CT. Specifically, the current estimate $\hat{x}$ obtained via DiffSolver is concatenated with the degraded XCT observation and processed through a Pix2Pix network. The use of degraded XCT observations is motivated by real-world scenarios, where scans of microstructures are corrupted by noise, blur, or insufficient number of views used for its reconstruction. This is one of the main advantages of our method, since it does not require clean auxiliary information, and still captures the essential cross-modal information.

The network is trained to map a wide range of acquisition conditions---including variations in noise, blur, number of views, and sparsity levels---to an ``ideal'' reference scenario characterized by minimal noise and blur, 512 views and dense sampling as shown in Fig.~\ref{fig:100grid}. This design serves two complementary purposes. First, it enforces consistency across modalities by aligning the evolving NCT reconstruction with corresponding XCT information. Second, it removes unknown artifacts that lack an explicit forward model, such as sensor noise and motion-induced blur. By conditioning on both $\hat{x}$ and the auxiliary modality, the network learns to exploit cross-modal redundancies and produces more informative reconstructions.

\noindent

\section{Experimental Results}
\label{sec:experiments}

\begin{figure}[t]
\centering
% First row: XCT images
\begin{subfigure}[b]{0.10\textwidth}
    \centering
    \includegraphics[width=\textwidth]{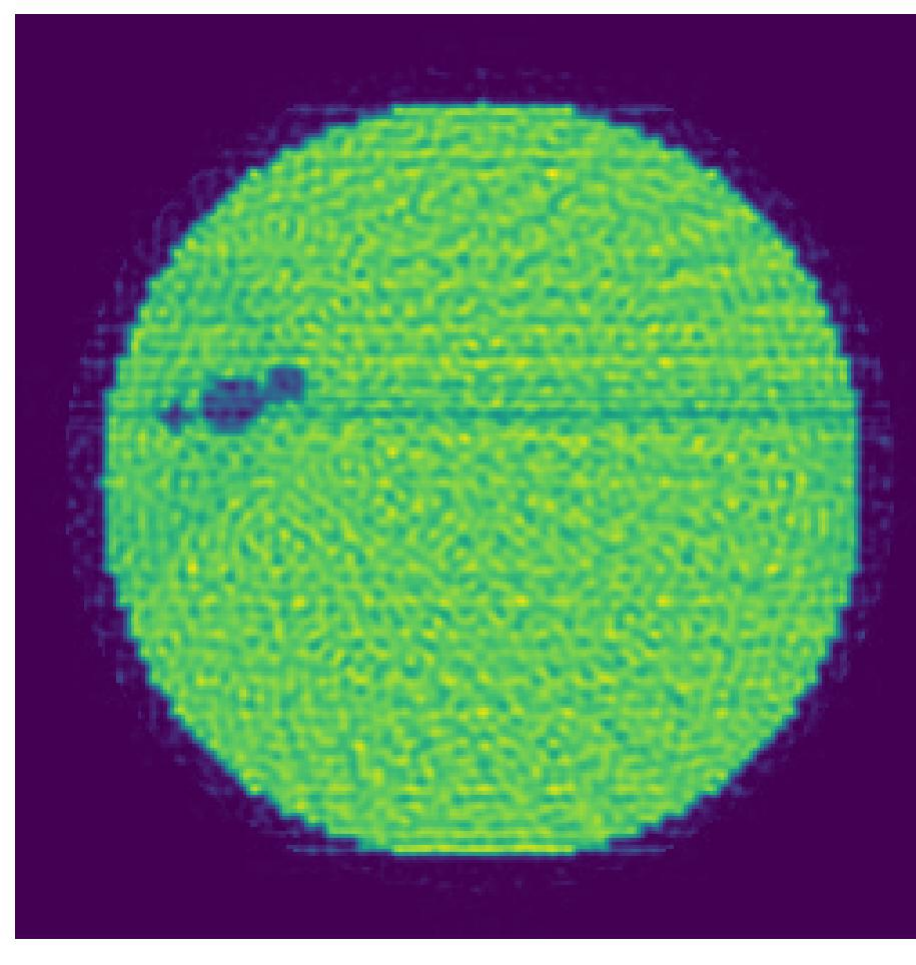}
    
\end{subfigure}
\begin{subfigure}[b]{0.10\textwidth}
    \centering
    \includegraphics[width=\textwidth]{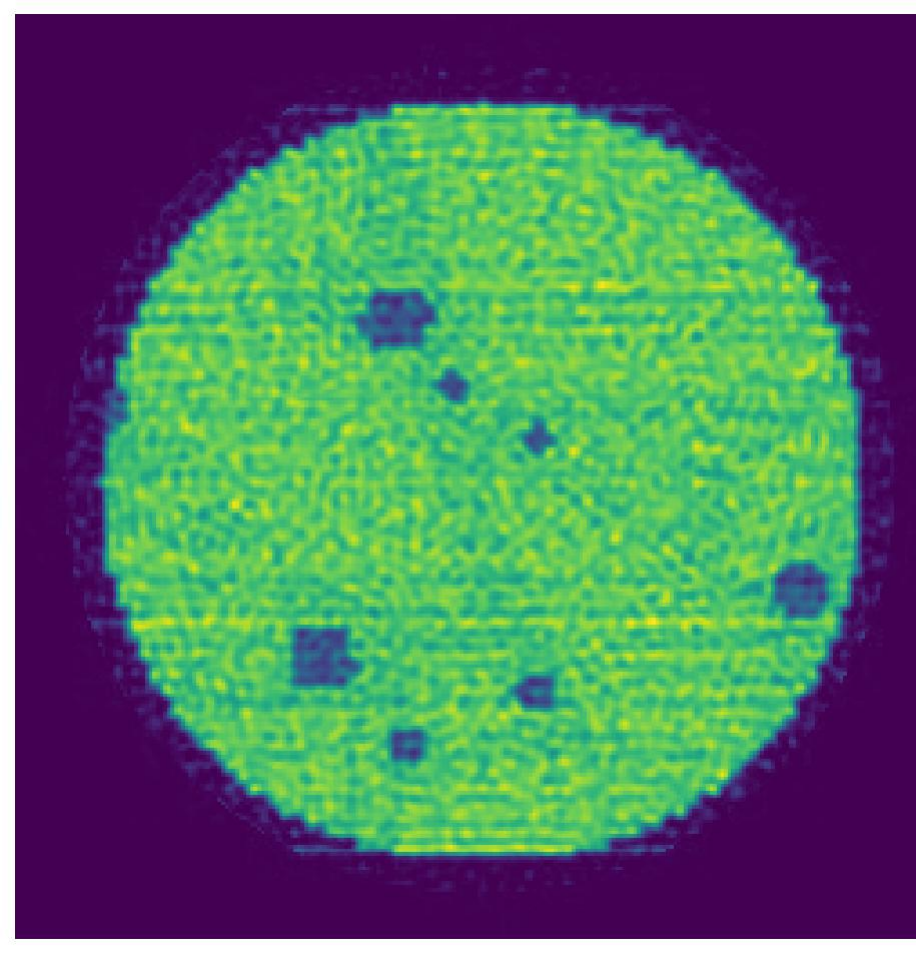}
    
\end{subfigure}
\begin{subfigure}[b]{0.10\textwidth}
    \centering
    \includegraphics[width=\textwidth]{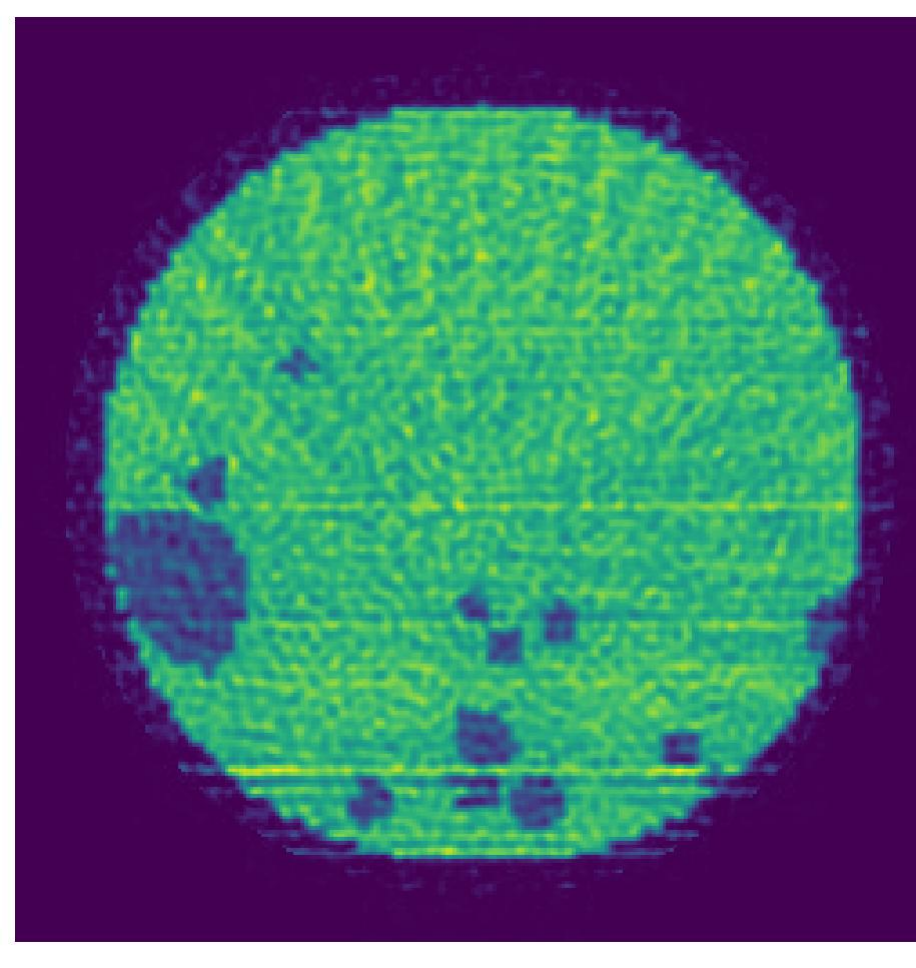}
    
\end{subfigure}
\begin{subfigure}[b]{0.10\textwidth}
    \centering
    \includegraphics[width=\textwidth]{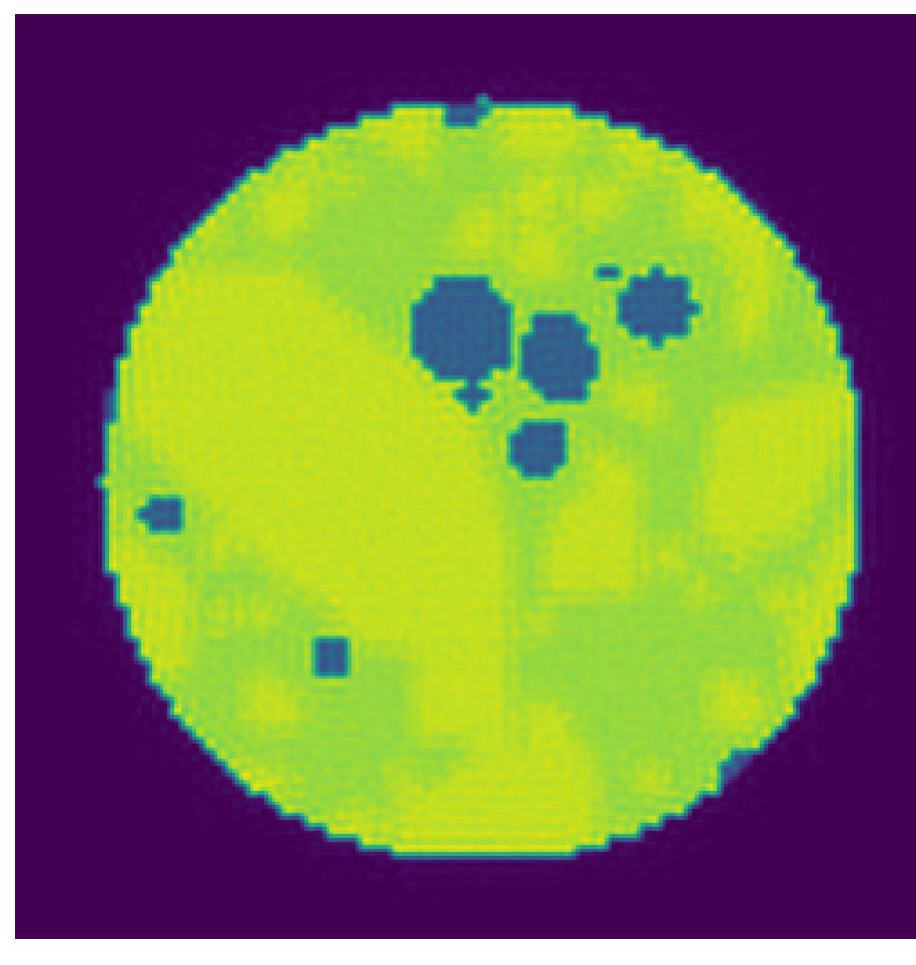}
    
\end{subfigure}

\vspace{2mm}
% Second row: Labels
\begin{subfigure}[b]{0.10\textwidth}
    \centering
    \includegraphics[width=\textwidth]{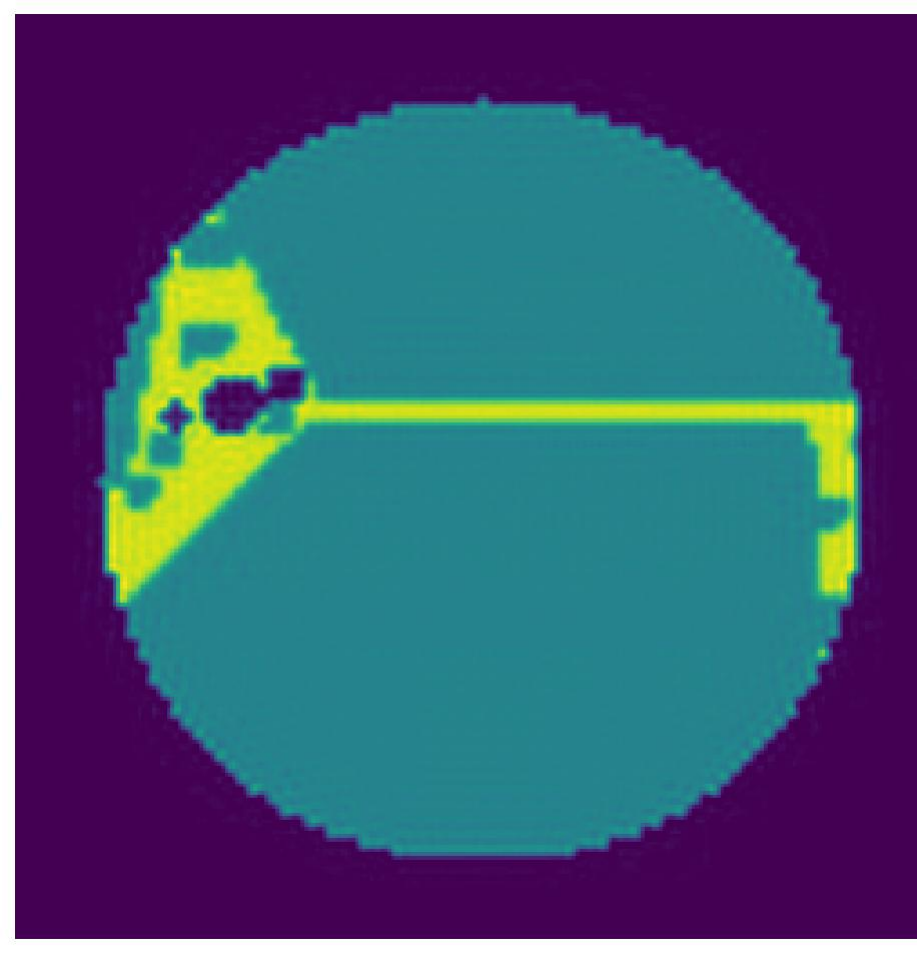}
\end{subfigure}
\begin{subfigure}[b]{0.10\textwidth}
    \centering
    \includegraphics[width=\textwidth]{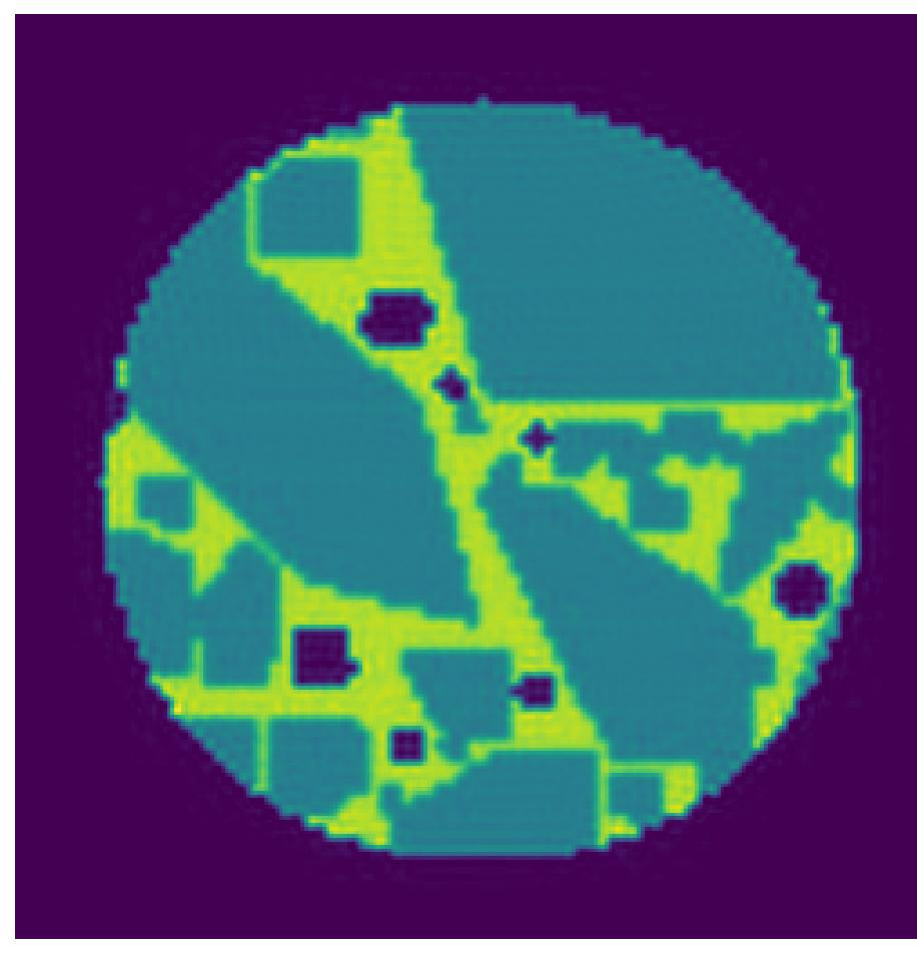}
\end{subfigure}
\begin{subfigure}[b]{0.10\textwidth}
    \centering
    \includegraphics[width=\textwidth]{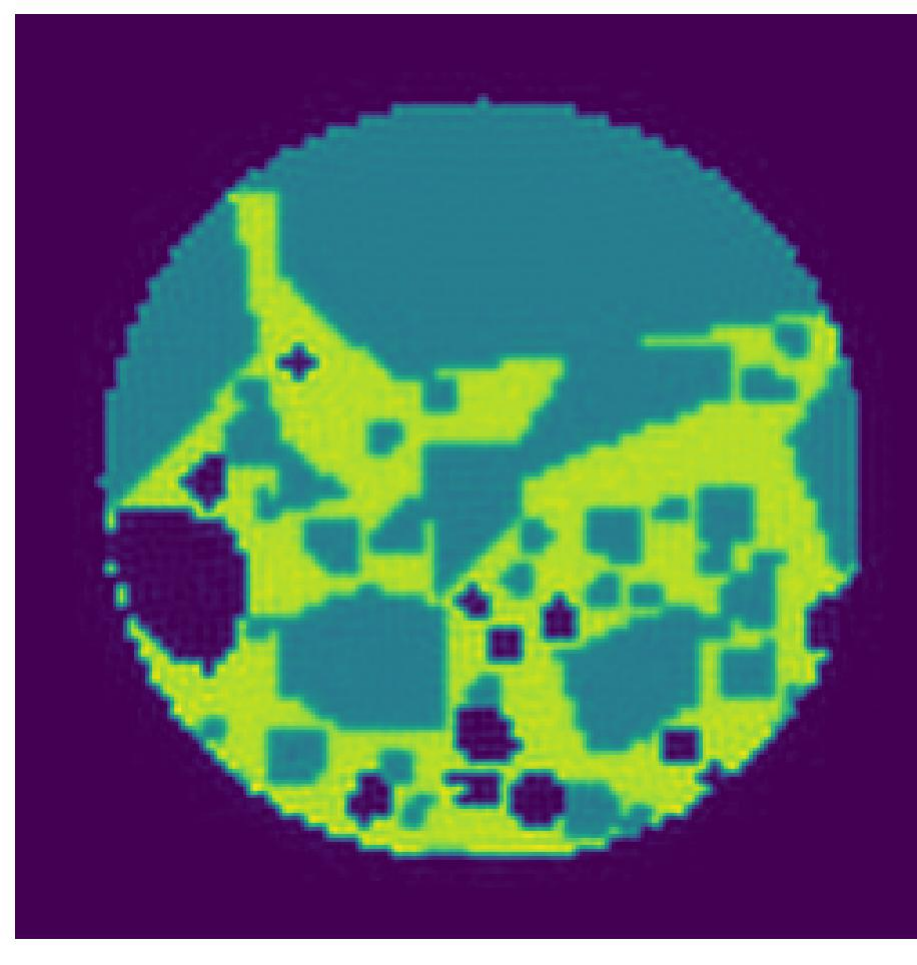}
\end{subfigure}
\begin{subfigure}[b]{0.10\textwidth}
    \centering
    \includegraphics[width=\textwidth]{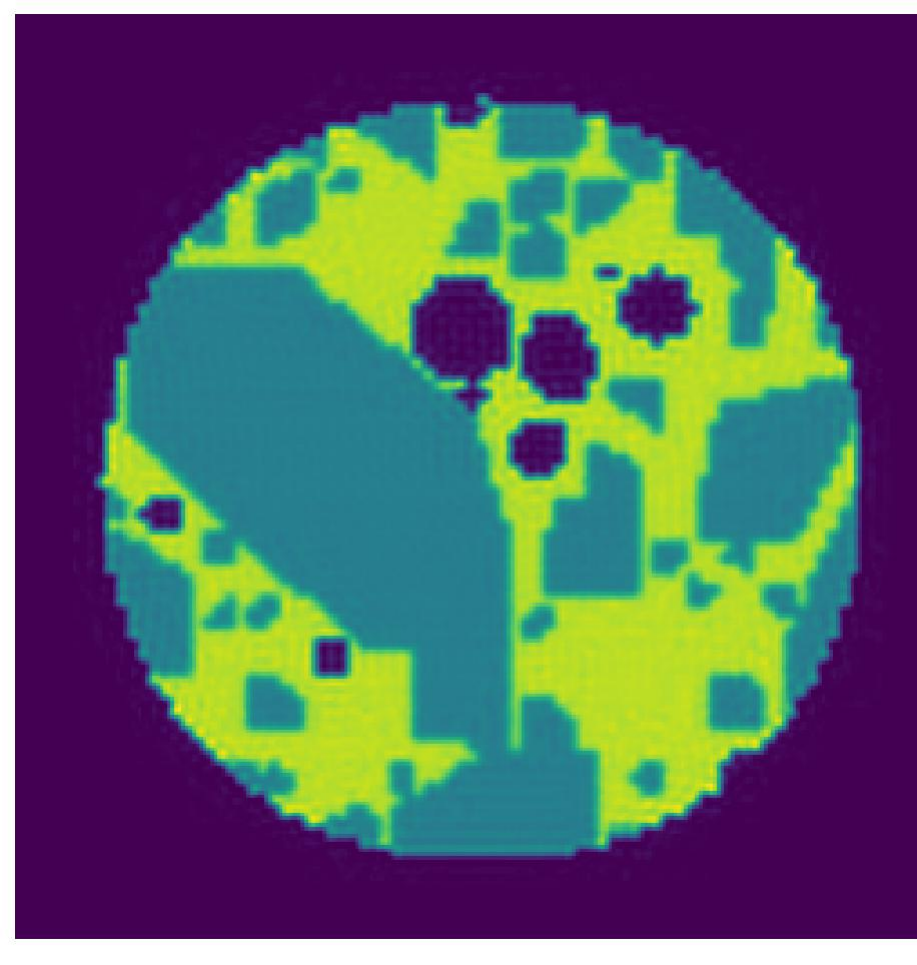}
\end{subfigure}

\caption{Comparison between XCT images (top row) and corresponding ideal NCT images (bottom row). XCT quality varies widely relative to NCT.}
\label{fig:xct_vs_label}
\end{figure}

\subsection{Dataset Collection and Experimental Setup}

We evaluated our framework on simulated cross-modal CT data, where neutron CT (NCT) and X-ray CT (XCT) projections of microstructure data~\cite{ziabari2025pycmg} are generated under controlled forward models with parallel-beam geometry. 10 unique 256$\times$256$\times$256 3D volumes are used for training, and 3 separate volumes with different microstructures for testing. 
For each volume, we simulate reconstructions under varying degradations, including noise, blur, sparsity levels, and number of views. We pair each degraded NCT/XCT sample with its label (ideal reconstruction) and train the Pix2Pix image-to-image translation networks \cite{isola2017image}. The network learns the consistent the features between NCT and XCT and removes the degradation artifacts, as shown in Fig.~\ref{fig:100grid}. The degraded simulations for NCT allow us to plug in intermediate reconstructions in the algorithm at every iteration and integrate the cross-modal information.
All experiments were performed on NVIDIA H100 GPUs. The execution time of the Pix2Pix model is negligible, contributing less than 1\% of total reconstruction time.

\subsection{Reconstruction Results}

Examples of XCT guidance samples and corresponding ideal NCT labels are shown in Fig.~\ref{fig:xct_vs_label}. 
XCT quality varies widely relative to NCT, and some information is modality-specific. Nevertheless, XCT provides structural cues that guide sparse-view NCT reconstruction.

Table~\ref{tab:psnr-ssim-steps} summarizes the reconstruction performance of unimodal (i.e., D3IP, with diffusion prior and without using second modality) and cross-modal (i.e., proposed with second auxiliary modality) approaches for 10 alternating iterations of prior enforcement and data consistency, under different numbers of optimization steps and projection views, with no measurement noise in the projections. The results are reported in terms of the Peak Signal-to-Noise Ratio (PSNR) and Structural Similarity Index Measure (SSIM), with the best score in each comparison highlighted in bold.

The main quantitative results are reported in Table~\ref{tab:psnr-ssim-steps}. Across most settings and numbers of views, and in particular in the sparse-view case, the cross-modal approach outperforms the baseline D3IP in terms of both PSNR and SSIM. In the \emph{low-view regime} ($8$--$32$ views), cross-modal consistency yields the largest gains, with improvements of up to $+1.63$ dB in PSNR and $+0.13$ in SSIM (5 steps, 32 views). These results indicate that cross-modal guidance is especially valuable when projection data is scarce. 
In \emph{high-view regime} ($128$--$256$ views), improvements in PSNR are smaller and occasionally slightly negative (e.g., $-0.20$ dB at 5 steps, 128 views), but SSIM consistently improves, with increases of up to $+0.15$ (5 and 10 steps, 256 views). This shows that even when reconstructions are already strong, the cross-modal block enhances structural fidelity and perceptual sharpness, highlighting density differences in the microstructure more clearly.

Under $5\%$ Gaussian measurement noise, the cross-modal method consistently outperforms the baseline D3IP, with average improvements of about $+0.5$ dB in PSNR and $+0.02$ in SSIM across settings. This demonstrates robustness to noise and confirms that the model can effectively integrate information from neutron CT projections and auxiliary XCT references even in degraded measurement conditions.

\begin{table}[!h]
\centering
\caption{Comparison of reconstruction quality (PSNR [dB] / SSIM) for Unimodal vs. Cross-modal for different numbers of views, no measurement noise, and at $T=10$ alternating steps between enforcing the prior and data consistency. 
$\Delta=\text{Cross}-\text{Uni}$.}
\label{tab:psnr-ssim-steps}
\resizebox{\columnwidth}{!}{%
\begin{tabular}{c|c|cc|cc|cc}
\toprule
\multirow{2}{*}{Steps} & \multirow{2}{*}{\#Views} & \multicolumn{2}{c|}{Unimodal} & \multicolumn{2}{c|}{Cross-modal} & \multicolumn{2}{c}{$\Delta$} \\
\cmidrule(lr){3-4}\cmidrule(lr){5-6}\cmidrule(l){7-8}
 &  & PSNR $\uparrow$ & SSIM $\uparrow$ & PSNR $\uparrow$ & SSIM $\uparrow$ & PSNR & SSIM \\
\midrule
\multirow{6}{*}{5} 
 & 8   & 20.10 & 0.374 & \textbf{21.17} & \textbf{0.375} & \textbf{+1.07} & \textbf{+0.001} \\
 & 16  & 23.34 & 0.455 & \textbf{23.97} & \textbf{0.471} & \textbf{+0.63} & \textbf{+0.016} \\
 & 32  & 25.82 & 0.490 & \textbf{27.45} & \textbf{0.620} & \textbf{+1.63} & \textbf{+0.130} \\
 & 64  & 28.62 & 0.648 & \textbf{29.12} & \textbf{0.762} & \textbf{+0.50} & \textbf{+0.114} \\
 & 128 & \textbf{28.74} & 0.662 & 28.54 & \textbf{0.777} & -0.20 & \textbf{+0.115} \\
 & 256 & 28.53 & 0.658 & \textbf{29.50} & \textbf{0.812} & \textbf{+0.97} & \textbf{+0.154} \\
\midrule
\multirow{6}{*}{10} 
 & 8   & 20.33 & 0.381 & \textbf{21.73} & \textbf{0.405} & \textbf{+1.40} & \textbf{+0.024} \\
 & 16  & 23.23 & 0.451 & \textbf{24.73} & \textbf{0.510} & \textbf{+1.50} & \textbf{+0.059} \\
 & 32  & 27.01 & 0.557 & \textbf{28.11} & \textbf{0.636} & \textbf{+1.10} & \textbf{+0.079} \\
 & 64  & \textbf{29.16} & 0.694 & 28.90 & \textbf{0.745} & -0.26 & \textbf{+0.051} \\
 & 128 & 29.37 & 0.730 & \textbf{29.73} & \textbf{0.818} & \textbf{+0.36} & \textbf{+0.088} \\
 & 256 & 28.58 & 0.657 & \textbf{29.30} & \textbf{0.811} & \textbf{+0.72} & \textbf{+0.154} \\
\midrule
\multirow{6}{*}{20} 
 & 8   & 20.54 & 0.381 & \textbf{21.87} & \textbf{0.422} & \textbf{+1.33} & \textbf{+0.041} \\
 & 16  & 23.44 & 0.466 & \textbf{24.79} & \textbf{0.508} & \textbf{+1.35} & \textbf{+0.042} \\
 & 32  & \textbf{26.85} & 0.558 & 26.93 & \textbf{0.599} & \textbf{+0.08} & \textbf{+0.041} \\
 & 64  & 28.85 & 0.665 & \textbf{29.56} & \textbf{0.772} & \textbf{+0.71} & \textbf{+0.107} \\
 & 128 & 29.83 & 0.759 & \textbf{29.97} & \textbf{0.816} & \textbf{+0.14} & \textbf{+0.057} \\
 & 256 & 30.07 & 0.768 & \textbf{30.21} & \textbf{0.812} & \textbf{+0.14} & \textbf{+0.044} \\
\bottomrule
\end{tabular}%
}
\end{table}

\begin{figure}[!h]
\centering
% First row: 8 views
\begin{subfigure}[b]{0.15\textwidth}
    \centering
    \includegraphics[width=\textwidth]{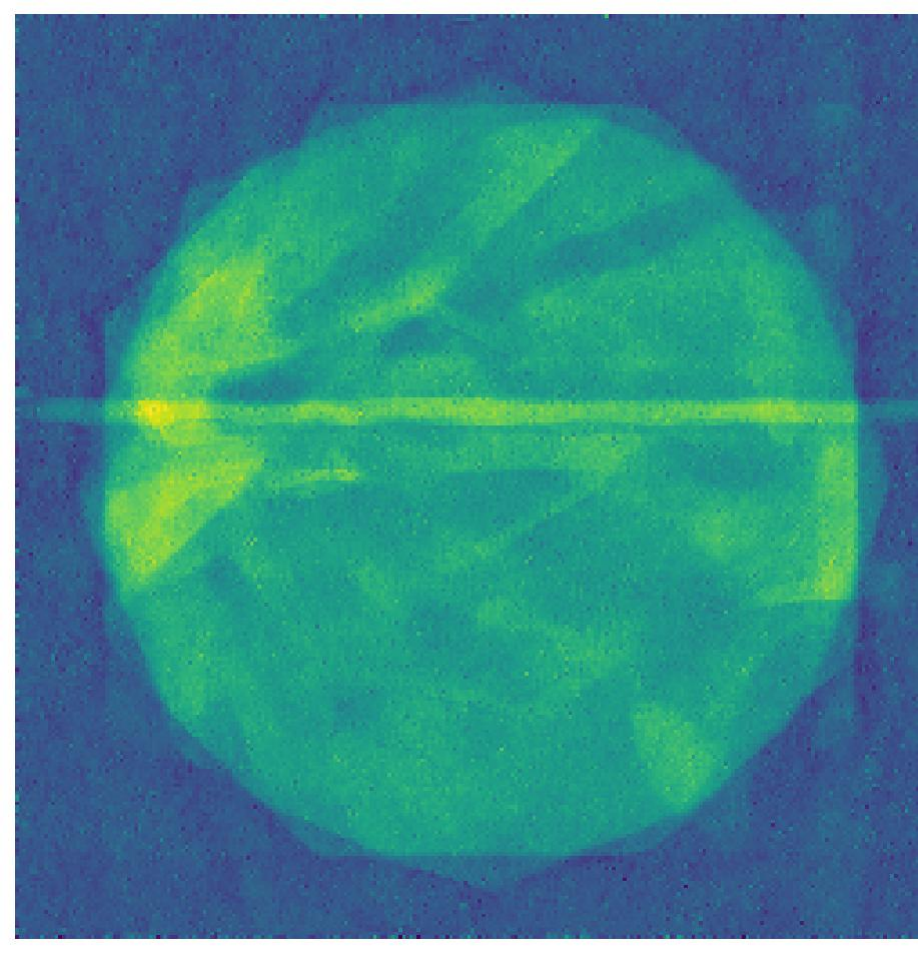}
    \caption*{Unimodal (D3IP)}
\end{subfigure}
\hfill
\begin{subfigure}[b]{0.15\textwidth}
    \centering
    \includegraphics[width=\textwidth]{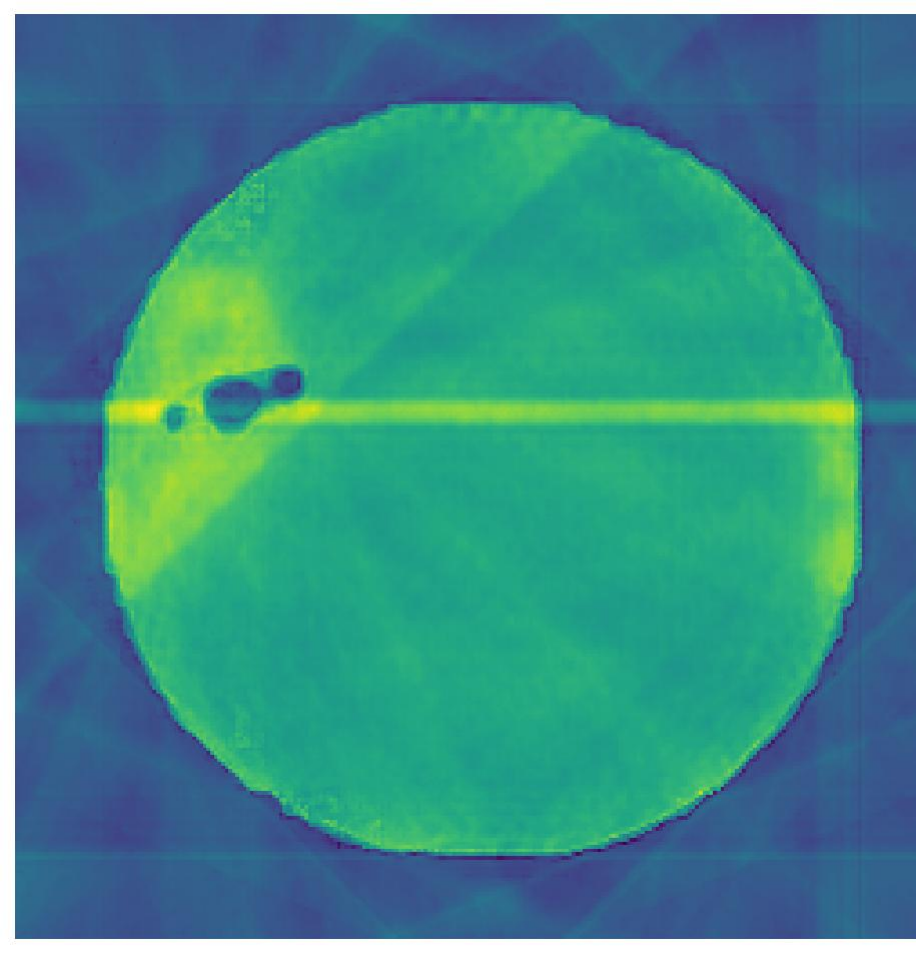}
    \caption*{Cross-modal (ours)}
\end{subfigure}
\hfill
\begin{subfigure}[b]{0.15\textwidth}
    \centering
    \includegraphics[width=\textwidth]{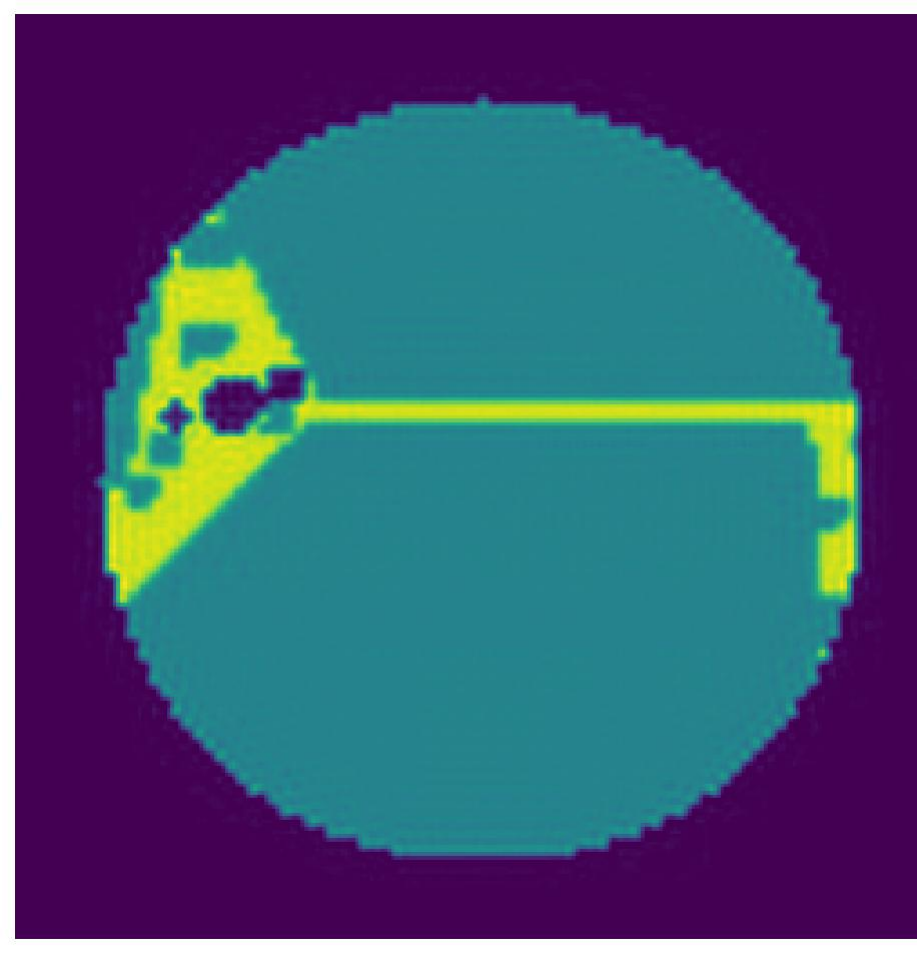}
    \caption*{Ground truth sample}
\end{subfigure}

% Second row: 16 views
\begin{subfigure}[b]{0.15\textwidth}
    \centering
    \includegraphics[width=\textwidth]{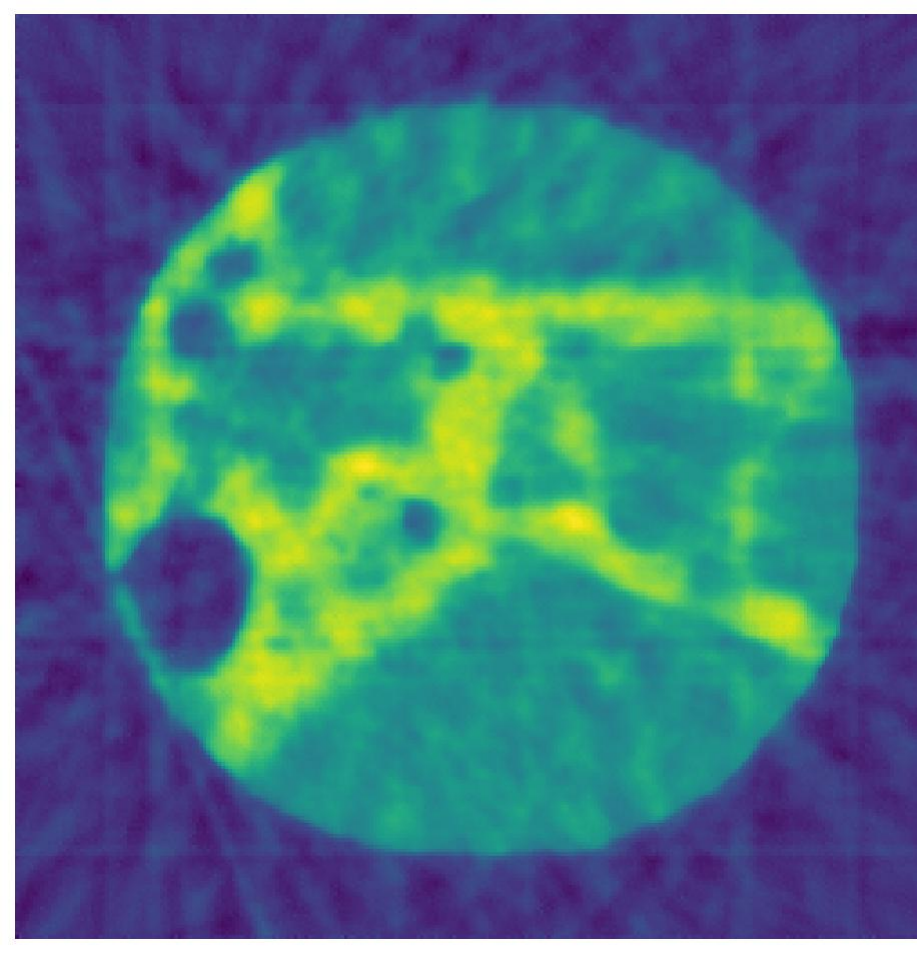}
    \caption*{Unimodal (D3IP)}
\end{subfigure}
\hfill
\begin{subfigure}[b]{0.15\textwidth}
    \centering
    \includegraphics[width=\textwidth]{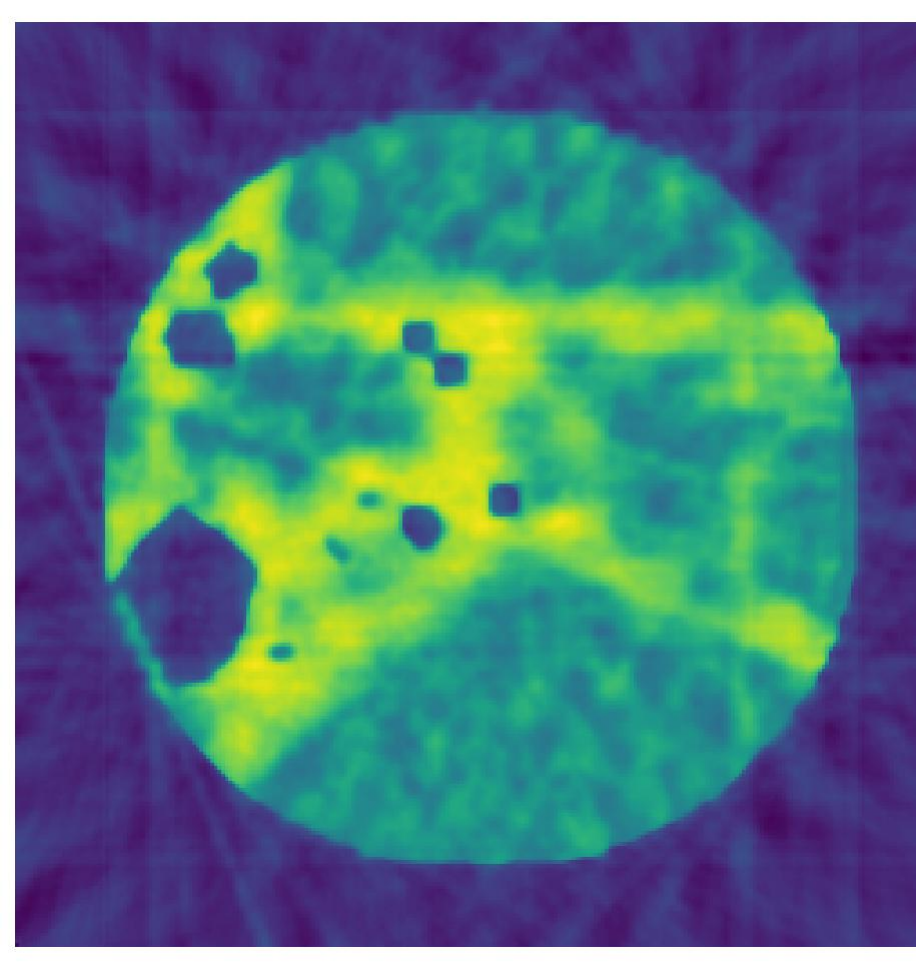}
    \caption*{Cross-modal (ours)}
\end{subfigure}
\hfill
\begin{subfigure}[b]{0.15\textwidth}
    \centering
    \includegraphics[width=\textwidth]{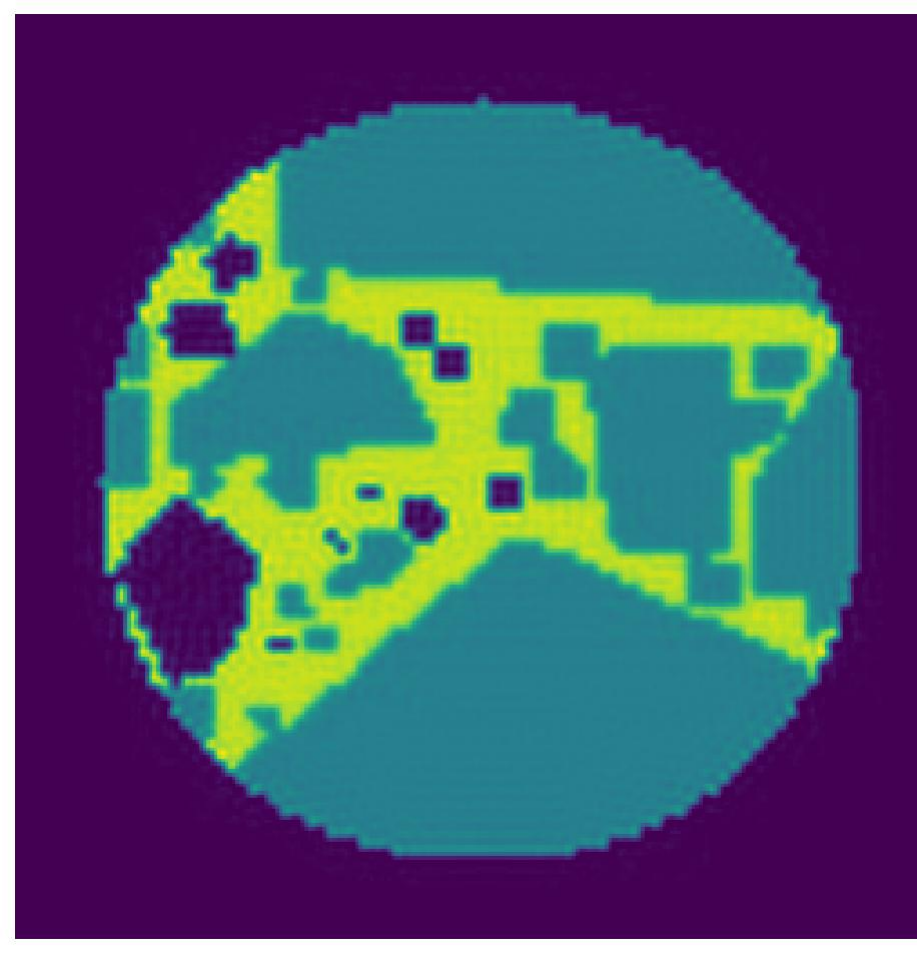}
    \caption*{Ground truth sample}
\end{subfigure}

% Third row: 32 views
\begin{subfigure}[b]{0.15\textwidth}
    \centering
    \includegraphics[width=\textwidth]{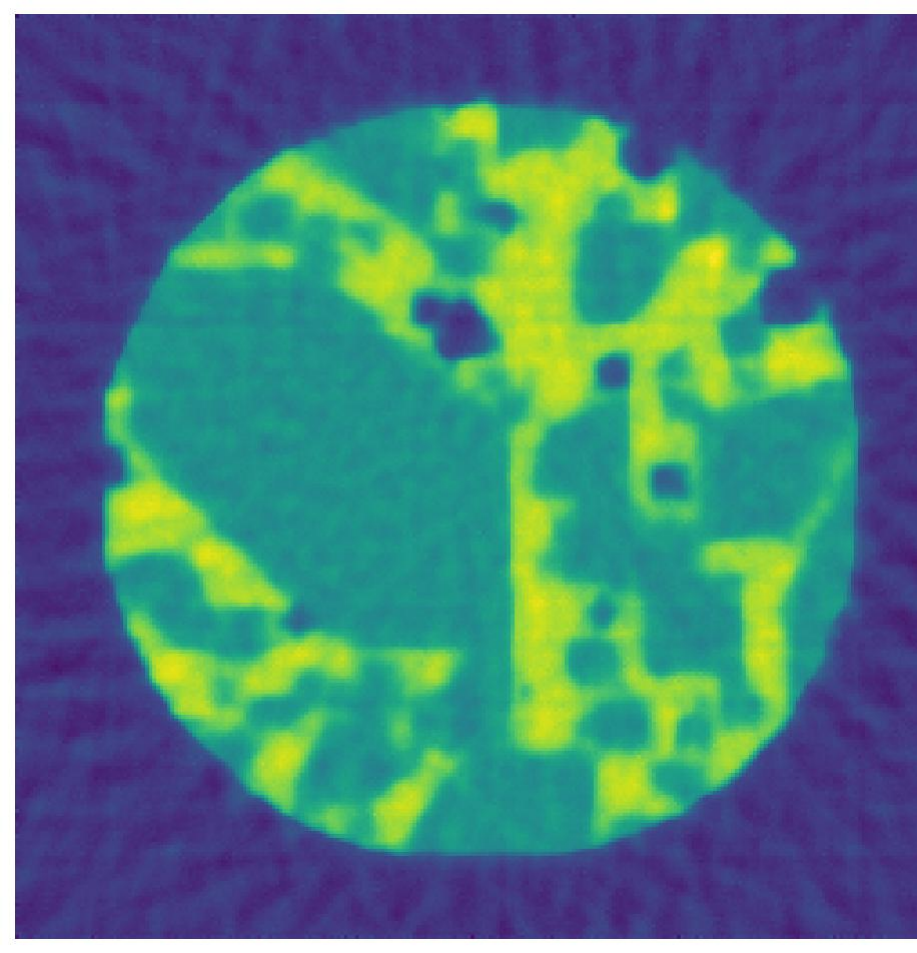}
    \caption*{Unimodal (D3IP)}
\end{subfigure}
\hfill
\begin{subfigure}[b]{0.15\textwidth}
    \centering
    \includegraphics[width=\textwidth]{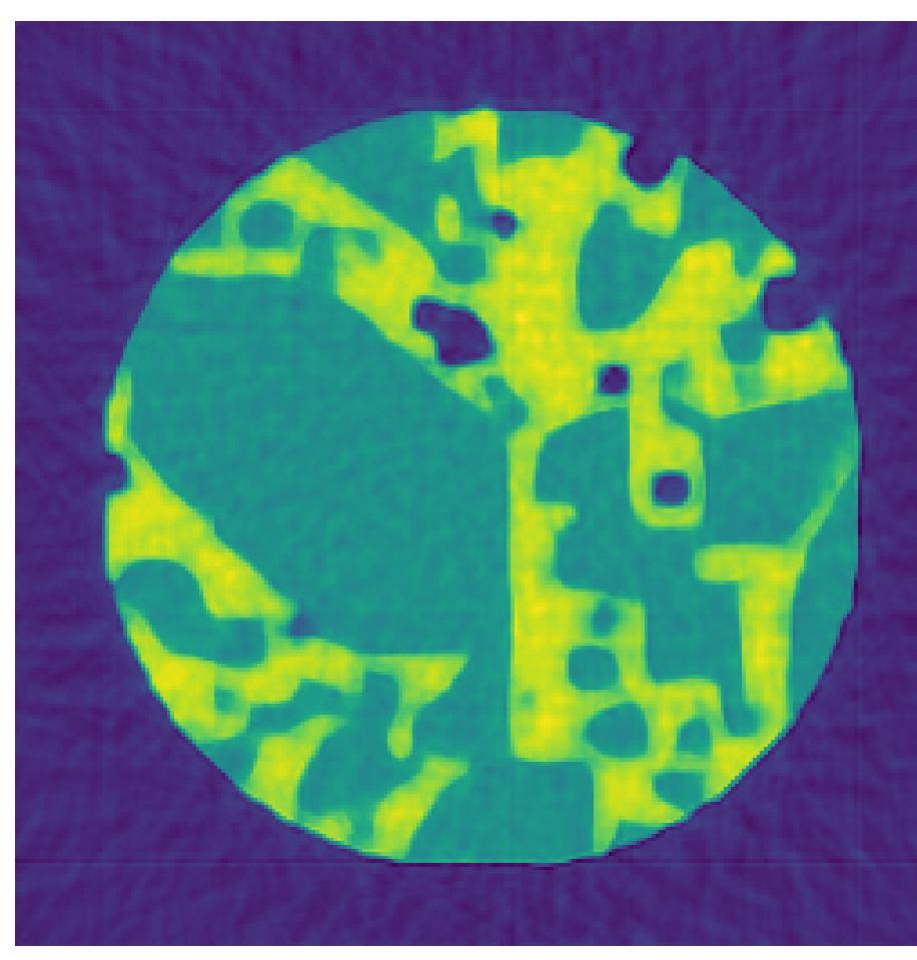}
    \caption*{Cross-modal (ours)}
\end{subfigure}
\hfill
\begin{subfigure}[b]{0.15\textwidth}
    \centering
    \includegraphics[width=\textwidth]{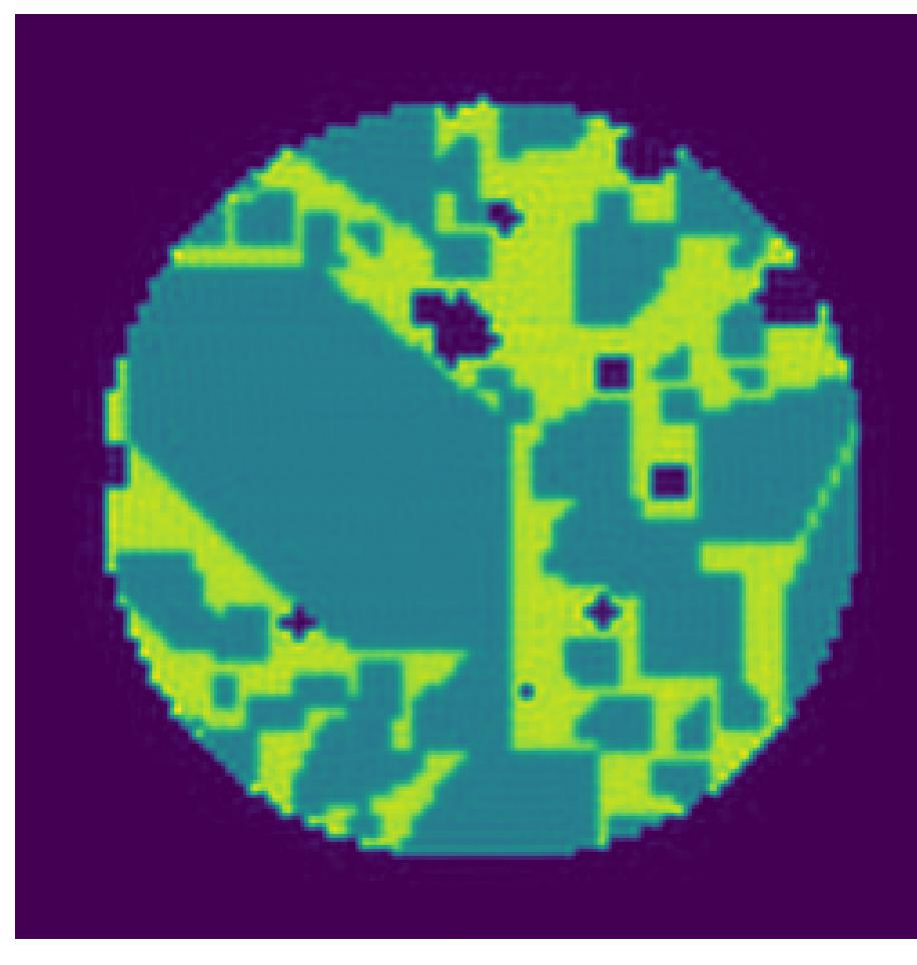}
    \caption*{Ground truth sample}
\end{subfigure}

% Fourth row: 64 views
\begin{subfigure}[b]{0.15\textwidth}
    \centering
    \includegraphics[width=\textwidth]{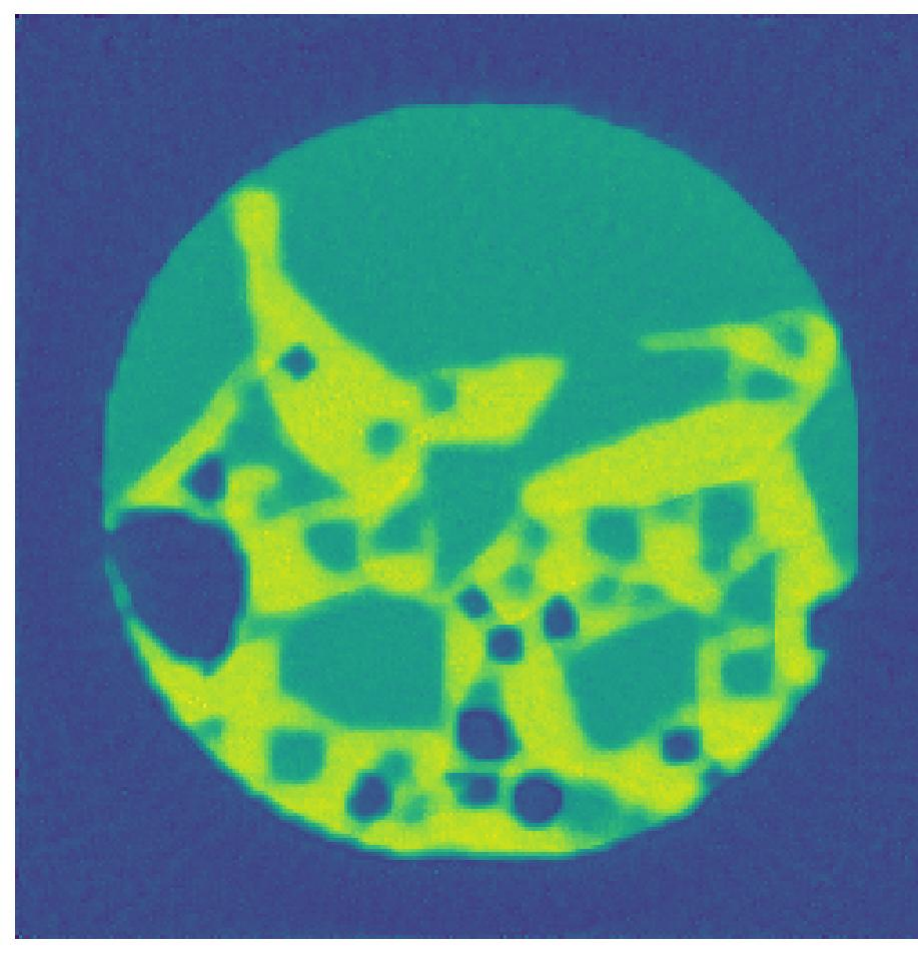}
    \caption*{Unimodal (D3IP)}
\end{subfigure}
\hfill
\begin{subfigure}[b]{0.15\textwidth}
    \centering
    \includegraphics[width=\textwidth]{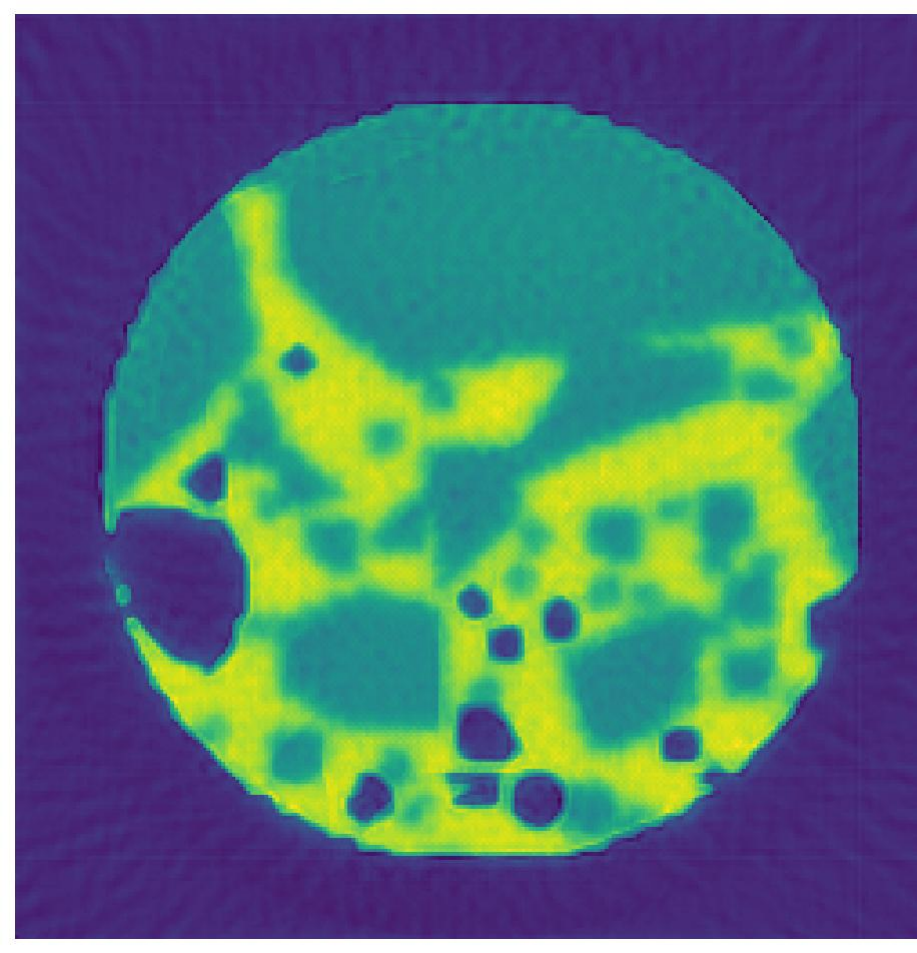}
    \caption*{Cross-modal (ours)}
\end{subfigure}
\hfill
\begin{subfigure}[b]{0.15\textwidth}
    \centering
    \includegraphics[width=\textwidth]{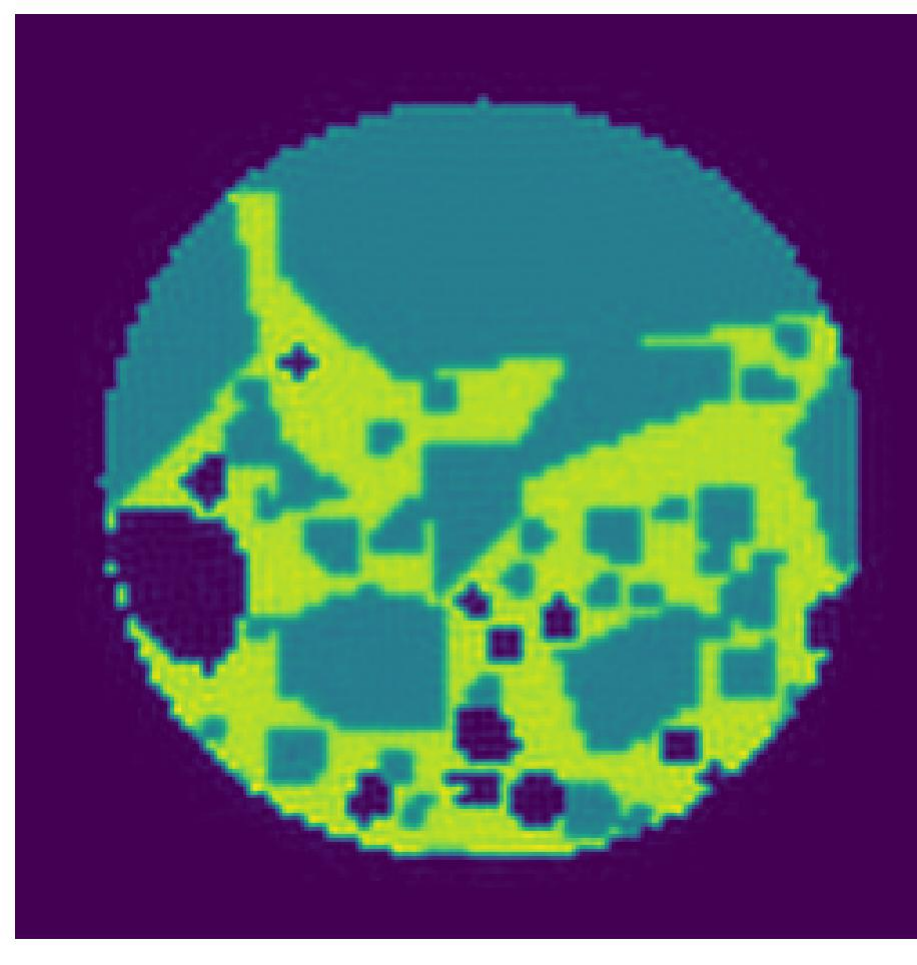}
    \caption*{Ground truth sample}
\end{subfigure}

\caption{Qualitative comparison between D3IP and our cross-modal guided algorithm across different numbers of projection views ($8$, $16$, $32$, and $64$ views, in rows 1–4, respectively). 
Within each row, the first column shows Unimodal (D3IP) reconstructions, the second column shows our cross-modal guided algorithm reconstructions, and the third column shows the ground truth samples.}
\label{fig:recon_views}
\end{figure}

In Fig.~\ref{fig:recon_views}, we compare reconstructions for 8, 16, 32, and 64 views in the sparse-view setting. With only 8 views, cross-modal reconstruction guided by XCT better recovers shapes and boundaries, closely matching the ground truth and capturing small features (e.g., the dark-blue region) with higher fidelity. Similar improvements appear for 16–64 views, where cross-modal consistency sharpens boundaries, reduces blur, and improves reconstruction of low-density regions. Overall, cross-modal reconstructions achieve higher visual fidelity than unimodal ones, particularly in sparse-view regimes where XCT guidance is most valuable.

% !TEX root = ../main.tex
\section{Conclusion}
\label{sec:conclusion}

In this work, we presented an algorithm to guide image reconstruction with diffusion priors and cross-modal information. 
Specifically, this algorithm introduces a lightweight cross-modal consistency module without retraining the diffusion model, and generalizes well to corrupted cross-modal information. 
We showed that the improvement provided by the algorithm is twofold: it both improves the intermediate and final reconstruction quality, but also more accurately fine-tunes the weights of the diffusion prior in the domain adaptation setting. 
The algorithm is validated on simulated sparse neutron CT data under the guidance of more easily attainable X-ray CT data. 
The X-ray CT acquisition setting was simulated under imperfect conditions (i.e., sparse/noisy) to demonstrate generalizability. 
On average, cross-modal guidance improved reconstructions by +0.73 dB PSNR and +0.073 SSIM, with the largest PSNR gains observed in the sparse-view regime (8–32 views). For future work, we plan to conduct experiments on real Neutron/X-ray CT pairs of data, and provide theoretical guarantees for cross-modal reconstruction.

\section{Acknowledgement}
\label{sec:ack}

The authors would like to thank Professor Yuejie Chi of Yale University for guidance on diffusion models and inverse problems.

\bibliographystyle{IEEEbib}
\bibliography{strings,refs}

@article{vlassenbroeck2007comparative,
  title={A comparative and critical study of X-ray CT and neutron CT as non-destructive material evaluation techniques},
  author={Vlassenbroeck, Jelle and Cnudde, Veerle and Masschaele, Bert and Dierick, Manuel and Van Hoorebeke, Luc and Jacobs, Patric},
  year={2007}
}

@misc{ziabari2025pycmg,
  author       = {Ziabari, Amir and Alnaggar, Mohamed and Cheniour, Amani},
  title        = {PyCMG-based Simulation of Volumetric Concrete Microstructure},
  year         = {2025},
  month        = apr,
  note         = {Dataset},
  institution  = {Oak Ridge National Laboratory},
  doi = {10.13139/ORNLNCCS/2548178},
  howpublished          = {https://doi.ccs.ornl.gov/dataset/4de41b4c-88d3-58cc-a6a5-07d3348a6576}
}

@inproceedings{gleason2010x,
  title={X-ray and neutron imaging for plant system biology investigations},
  author={Gleason, Shaun S and Paquit, Vincent C and Bilheux, Hassina Z and Willis, Keely J and Deleon, Alyssa M and McNuttz, WM and Kalluri, Udaya C},
  booktitle={Future of Instrumentation International Workshop},
  year={2010}
}

@article{kim2013high,
  title={High-resolution neutron and X-ray imaging of granular materials},
  author={Kim, Felix H and Penumadu, Dayakar and Gregor, Jens and Kardjilov, Nikolay and Manke, Ingo},
  journal={Journal of Geotechnical and Geoenvironmental Engineering},
  volume={139},
  number={5},
  pages={715--723},
  year={2013},
  publisher={American Society of Civil Engineers}
}

@article{venkatakrishnan2021convolutional,
  title={Convolutional neural network based non-iterative reconstruction for accelerating neutron tomography},
  author={Venkatakrishnan, Singanallur and Ziabari, Amirkoushyar and Hinkle, Jacob and Needham, Andrew W and Warren, Jeffrey M and Bilheux, Hassina Z},
  journal={Machine Learning: Science and Technology},
  volume={2},
  number={2},
  pages={025031},
  year={2021},
  publisher={IOP Publishing}
}

@inproceedings{ho2020denoising,
  author    = {Jonathan Ho and Ajay Jain and Pieter Abbeel},
  title     = {Denoising Diffusion Probabilistic Models},
  booktitle = {NeurIPS},
  year      = {2020}
}

@article{song2019score,
  author  = {Yang Song and Stefano Ermon},
  title   = {Generative Modeling by Estimating Gradients of the Data Distribution},
  journal = {NeurIPS},
  year    = {2019}
}

@article{li2025cross,
  title={Cross-modal enhanced sparse CT imaging via null-space denoising diffusion with random medical measurement embedding},
  author={Li, X. and Shang, K. and Butala, M. D. and Wang, G.},
  journal={Alexandria Engineering Journal},
  volume={126},
  pages={565--577},
  year={2025},
  publisher={Elsevier}
}

@inproceedings{efimov2025leveraging,
  title={Leveraging Multimodal Diffusion Models to Accelerate Imaging with Side Information},
  author={Efimov, Timofey and Dong, Harry and Shah, Megna and Simmons, Jeff and Donegan, Sean and Chi, Yuejie},
  booktitle={ICASSP 2025-2025 IEEE International Conference on Acoustics, Speech and Signal Processing (ICASSP)},
  pages={1--5},
  year={2025},
  organization={IEEE}
}

@inproceedings{chung2023solving,
  title     = {Solving 3D Inverse Problems using Pre-trained 2D Diffusion Models},
  author    = {Chung, Hyungjin and Ryu, Dohyun and McCann, Michael T. and Klasky, Marc L. and Ye, Jong Chul},
  booktitle = {Proceedings of the IEEE/CVF Conference on Computer Vision and Pattern Recognition (CVPR)},
  pages     = {2586--2595},
  year      = {2023},
}

@inproceedings{chung2024deep,
  title     = {Deep Diffusion Image Prior for Efficient OOD Adaptation in 3D Inverse Problems},
  author    = {Chung, Hyungjin and Ye, Jong Chul},
  booktitle = {European Conference on Computer Vision (ECCV)},
  pages     = {432--455},
  year      = {2024},
  publisher = {Springer Nature Switzerland},
  address   = {Cham}
}

@inproceedings{kawar2022denoising,
  title     = {Denoising Diffusion Restoration Models},
  author    = {Kawar, Bahjat and Elad, Michael and Ermon, Stefano and Song, Jiaming},
  booktitle = {Advances in Neural Information Processing Systems (NeurIPS)},
  volume    = {35},
  pages     = {23593--23606},
  year      = {2022}
}

@inproceedings{xu2024provably,
  title     = {Provably Robust Score-Based Diffusion Posterior Sampling for Plug-and-Play Image Reconstruction},
  author    = {Xu, Xingyu and Chi, Yuejie},
  booktitle = {Advances in Neural Information Processing Systems (NeurIPS)},
  volume    = {37},
  pages     = {36148--36184},
  year      = {2024}
}

@inproceedings{nichol2021improved,
  title     = {Improved Denoising Diffusion Probabilistic Models},
  author    = {Nichol, Alexander Quinn and Dhariwal, Prafulla},
  booktitle = {Proceedings of the International Conference on Machine Learning (ICML)},
  pages     = {8162--8171},
  year      = {2021},
  organization = {PMLR}
}

@article{author202Xconversion,
  title={Conversion between CT and MRI images using diffusion and score-matching models},
  author={Lyu, Qiong and Wang, Ge},
  journal={arXiv preprint arXiv:2209.12104},
  year={2022}
}

@article{xia2025diffusion,
  title     = {A Diffusion Model Translator for Efficient Image-to-Image Translation},
  author    = {Xia, Mengfei and Zhou, Yu and Yi, Ran and Liu, Yong-Jin and Wang, Wenping},
  journal   = {IEEE Transactions on Pattern Analysis and Machine Intelligence},
  year      = {2024},
  volume    = {PP},
  number    = {99},
  pages     = {},
  doi       = {10.1109/TPAMI.2024.3435448},
  eprint    = {arXiv:2502.00307},
  note      = {Published online Feb 2025; early access}
}

@article{chung2023prompt,
  title   = {Prompt-Tuning Latent Diffusion Models for Inverse Problems},
  author  = {Chung, Hyungjin and Ye, Jong Chul and Milanfar, Peyman and Delbracio, Mauricio},
  journal = {arXiv preprint arXiv:2310.01110},
  year    = {2023}
}

@inproceedings{dou2024diffusion,
  title     = {Diffusion Posterior Sampling for Linear Inverse Problem Solving: A Filtering Perspective},
  author    = {Dou, Ziluo and Song, Yang},
  booktitle = {Proceedings of the Twelfth International Conference on Learning Representations (ICLR)},
  year      = {2024}
}

@inproceedings{isola2017image,
  title={Image-to-image translation with conditional adversarial networks},
  author={Isola, Phillip and Zhu, Jun-Yan and Zhou, Tinghui and Efros, Alexei A.},
  booktitle={Proceedings of the IEEE conference on computer vision and pattern recognition},
  pages={1125--1134},
  year={2017}
}

\end{document}